\documentclass[10pt,twocolumn,letterpaper]{article}

\usepackage[pagenumbers]{iccv} %

\definecolor{iccvblue}{rgb}{0.21,0.49,0.74}
\usepackage[pagebackref,breaklinks,colorlinks,allcolors=iccvblue]{hyperref}
\usepackage{tikz}
\usepackage{pgfplots}
\usepackage{comment}
\usepackage{multirow}
\usetikzlibrary{pie}
\usetikzlibrary{calc}
\usepgfplotslibrary{groupplots}

\pgfplotscreateplotcyclelist{scicolors}{
  {rgb={0.267,0.467,0.667},mark=o}\\     %
  {rgb={0.933,0.400,0.467},mark=square}\\ %
  {rgb={0.800,0.733,0.267},mark=diamond}\\ %
  {rgb={0.060,0.350,0.100},mark=triangle}\\ %
  {rgb={0.400,0.800,0.933},mark=pentagon}\\ %
  {rgb={0.667,0.200,0.467},mark=*}\\     %
}

\def\paperID{6195} %
\def\confName{ICCV}
\def\confYear{2025}

\title{Fast Autoregressive Models for Continuous Latent Generation}

\author{Tiankai Hang\thanks{Work done during internship at Microsoft Research Asia.} \quad Jianmin Bao \quad Fangyun Wei \quad Dong Chen\\
Microsoft Research Asia \\
}

\begin{document}
\maketitle
\begin{abstract}

Autoregressive models have demonstrated remarkable success in sequential data generation, particularly in NLP, but their extension to continuous-domain image generation presents significant challenges.
Recent work, the masked autoregressive model (MAR), bypasses quantization by modeling per-token distributions in continuous spaces using a diffusion head but suffers from slow inference due to the high computational cost of the iterative denoising process.
To address this, we propose the Fast AutoRegressive model (FAR), a novel framework that replaces MAR's diffusion head with a lightweight shortcut head, enabling efficient few-step sampling while preserving autoregressive principles. Additionally, FAR seamlessly integrates with causal Transformers, extending them from discrete to continuous token generation without requiring architectural modifications. 
Experiments demonstrate that FAR achieves $2.3\times$ faster inference than MAR while maintaining competitive FID and IS scores. This work establishes the first efficient autoregressive paradigm for high-fidelity continuous-space image generation, bridging the critical gap between quality and scalability in visual autoregressive modeling.

\end{abstract}
    
\section{Introduction}
\label{sec:intro}

Autoregressive models, such as GPT~\cite{radford2019gpt2,brown2020gpt3}, have emerged as the dominant framework for generative modeling in natural language processing, owing to their strong capability in sequential prediction tasks. These models operate in a categorical, discrete space, naturally aligning with the structure of language. However, extending this success to continuous-domain content generation, such as images, typically requires discretization techniques. Approaches like VQ-VAE~\cite{van2017neural,VQVAE2} and RQ-VAE~\cite{lee2022autoregressive} are commonly used for this purpose but often introduce information loss and training instability during quantization.

\begin{figure}[!t]
    \centering    \includegraphics[width=0.99\linewidth]{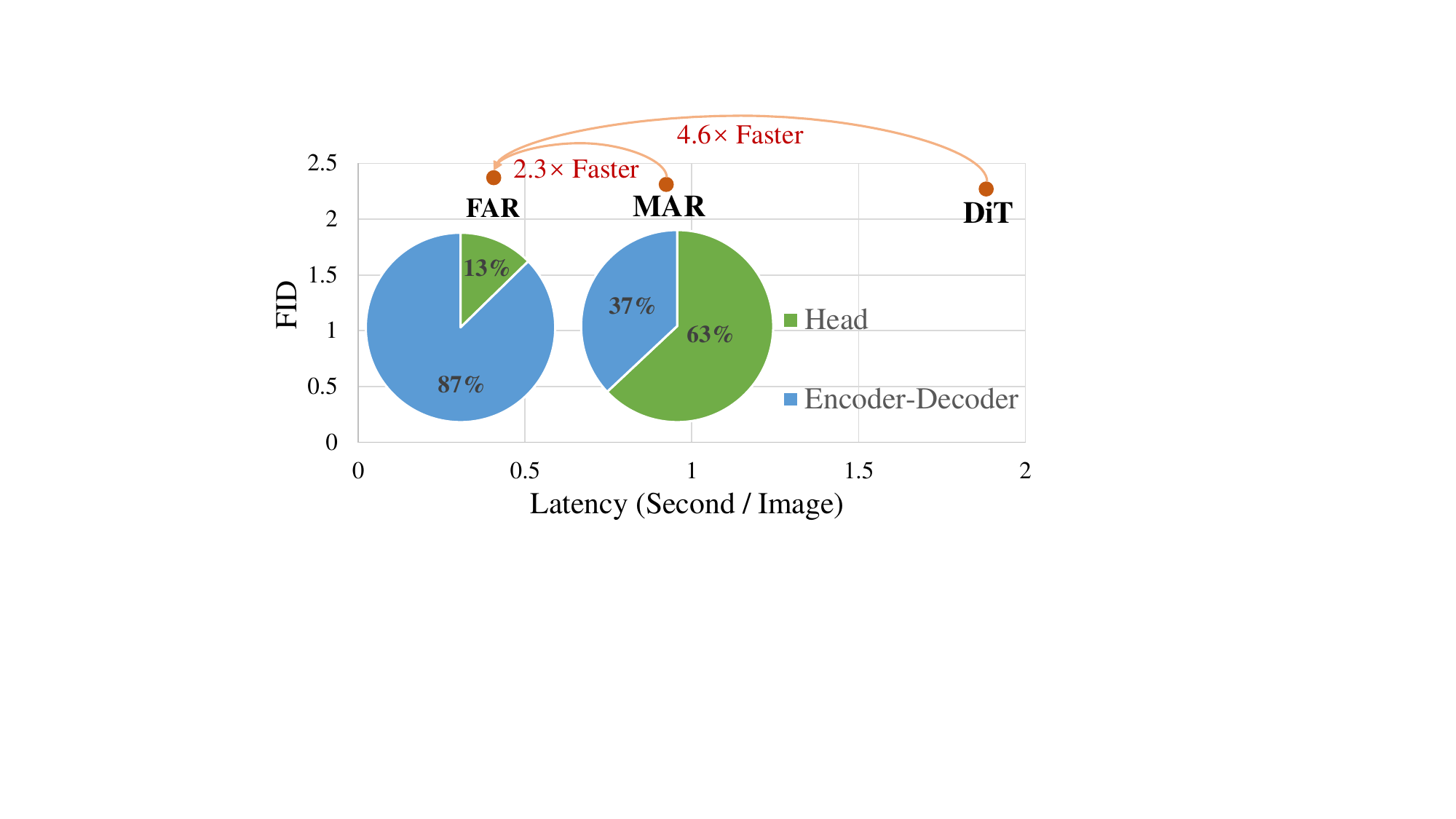}
    \caption{Inference cost breakdown and efficiency comparison among FAR, MAR~\cite{li2025mar}, and DiT~\cite{peebles2023dit} for  generating a $256 \times 256$ resolution image. Both FAR and MAR utilize the same encoder-decoder architecture, comprising 24 Transformer blocks with 172M parameters, along with a 6-layer MLP as the head network. In comparison, DiT (DiT-XL version), which achieves similar performance, features 28 Transformer blocks and 676M parameters. In MAR, the head network is the primary computational bottleneck, accounting for the majority of the inference cost. FAR mitigates this issue by introducing a more efficient head network that requires fewer denoising steps, achieving up to $2.3\times$ acceleration over MAR while maintaining nearly identical performance on ImageNet~\cite{deng2009imagenet} generation.}
    \label{fig:teaser-1}
\end{figure}

Recent research has shifted toward modeling per-token probabilistic distributions instead of discrete tokens to eliminate the need for image quantization, leveraging autoregressive principles.  Approaches like the masked autoregressive model (MAR)~\cite{li2025mar} exemplify this paradigm. MAR operates autoregressively, generating a set of image tokens using diffusion-based decoding, conditioned on all previously generated tokens at each iteration. Notably, each token resides in a continuous space defined by a pre-trained VAE. While MAR operates in continuous spaces, it suffers from slow inference due to the iterative nature of the diffusion process, often requiring hundreds of denoising steps per token generation. Figure~\ref{fig:teaser-1} illustrates the inference cost of each component in MAR, which comprises an encoder-decoder that generates conditions for the image tokens to be predicted, and a diffusion head that predicts the corresponding tokens through a multi-step denoising process. The inference cost analysis of MAR reveals that the head network contributes the majority of the total computation, accounting for an unexpected 63\% when generating an image with a resolution of $256 \times 256$, while the encoder-decoder consumes only 37\%. This is despite the head network being a simple 6-layer MLP, whereas the encoder-decoder is a substantially deeper and larger 24-layer Transformer.

\begin{figure}[!t]
    \centering
\includegraphics[width=0.99\linewidth]{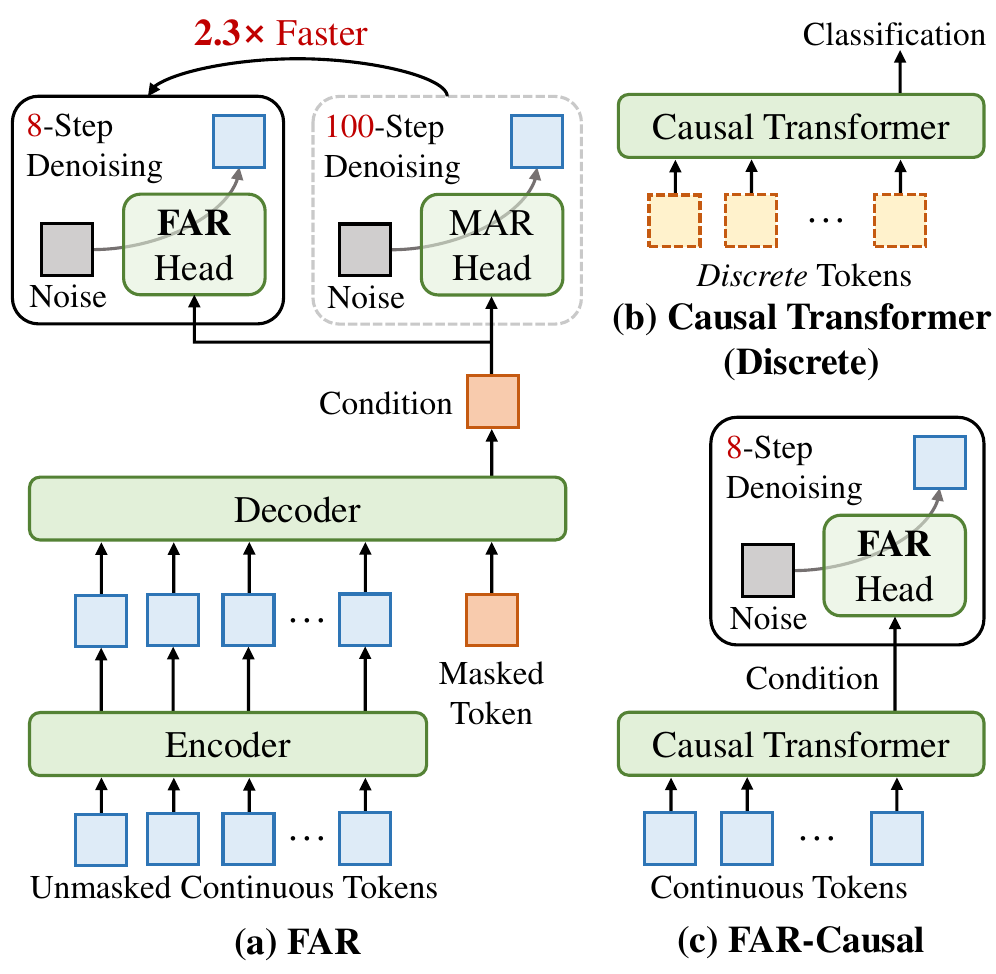}
    \caption{(a) FAR introduces a shortcut head that could replace the high-cost diffusion-based head in MAR, significantly reducing the inference cost while preserving the autoregressive principle and maintaining performance. (b-c) Integration of the FAR head enables a causal Transformer to transition from operating in a discrete space to a continuous space for image generation.}
    \label{fig:teaser-2}
\end{figure}

In this paper, we introduce \textbf{F}ast \textbf{A}uto\textbf{R}egressive model (FAR), a novel image generation model that preserves the autoregressive principle while significantly reducing the high inference cost of MAR~\cite{li2025mar}. FAR achieves this by incorporating a shortcut~\cite{frans2024one} head, which could seamlessly replace the high-cost diffusion-based head in MAR, as illustrated in Figure~\ref{fig:teaser-2}(a). Furthermore, this efficient head network can be directly integrated into a causal Transformer by substituting its original classification head with the FAR head, enabling the Transformer to transition from operating in a discrete space (Figure~\ref{fig:teaser-2}(b)) to a continuous space (Figure~\ref{fig:teaser-2}(c)) for image generation.

FAR offers several key advantages. First, it significantly reduces computational costs while maintaining output quality and preserving autoregressive principles, making it more practical for real-time applications compared to MAR~\cite{li2025mar}. Second, by modeling image latent distributions in continuous space, FAR eliminates the need for training discrete image tokenizers, which often require sophisticated optimization techniques~\cite{VQVAE2,VITVQGAN,zhu2024scaling,RQVAE} to mitigate challenges such as the limited expressiveness of fixed-size codebooks and training instability. Lastly, FAR enables causal Transformers to transition from operating in a discrete space to a continuous space for image generation. By tackling both the efficiency and complexity challenges, FAR marks a significant advancement in visual synthesis, enabling the development of more scalable and computationally efficient generative models for diverse continuous domains.

\section{Related Works}

\noindent\textbf{Autoregressive Models in Continuous Domains.}
Autoregressive models have been instrumental in generative tasks within natural language processing due to their inherent compatibility with discrete data structures~\cite{radford2018improving,brown2020language,achiam2023gpt,touvron2023llama,touvron2023llama2,dubey2024llama,team2023gemini,chowdhery2023palm,anil2023palm,bai2023qwen,yang2024qwen2,abdin2024phi}. However, their extension to continuous domains, such as images, has posed significant challenges. An early effort in this direction is iGPT~\cite{chen2020generative}, which is used for next-pixel prediction. Subsequently, discretization techniques like VQ-VAE~\cite{van2017neural} are introduced to compress images into discrete latent codes. Later models~\cite{yu2021vector,yu2022scaling,esser2021taming,VQVAE2} combine these learned tokens with transformer-based architectures to generate high-fidelity images. However, vector quantization introduces information loss and increases training sensitivity. These issues have sparked interest in developing models capable of directly handling continuous data.

Recently, there has been increasing interest in generative models for continuous token sequences. GIVT~\cite{tschannen2024givt} utilizes real-valued vector sequences for infinite-vocabulary synthesis, while HART~\cite{tang2024hart} integrates discrete-coarse and continuous-residual tokens to enable efficient 1024$\times$1024 image generation. JetFormer~\cite{tschannen2024jetformer} and DreamLLM~\cite{dong2023dreamllm} unify raw pixel-text autoregression to facilitate interleaved content creation. SEED-X~\cite{ge2024seed} bridges multi-granular comprehension and generation through shared representations, enhancing cross-modal consistency. A notable development is the integration of diffusion-based heads with masked autoregressive models (MAR)~\cite{li2025mar}, replacing discretization with diffusion-based probability modeling in continuous space. However, combining diffusion sampling with autoregressive sequence generation can lead to slow sampling speeds.

\noindent\textbf{Few-Step Diffusion Models.}
Recent advances in diffusion models focus on efficient sampling through high-order solvers and distillation. Efficient sampling methods~\cite{lu2022dpm,lu2022dpm++, karras2022elucidating, liu2022pseudo, zhao2023unipc} can dramatically reduce the number of sampling steps required by pre-trained diffusion models. The Analytic-DPM~\cite{bao2022analytic,bao2022estimating} established a theoretical foundation by analyzing discretization errors and convergence. Building on this, DPM-Solver~\cite{lu2022dpm} and DPM-Solver++~\cite{lu2022dpm++} leverage exact ODE solutions and data prediction for stable, high-quality sampling in just 10–20 steps, significantly enhancing latent-space models like Stable Diffusion.

Distillation methods~\cite{song2023consistency, liu2022flow, salimans2022progressive, meng2023distillation, gu2023boot, berthelot2023tract, zheng2022fast, luhman2021knowledge, xiao2021tackling} further accelerate sampling. For example, Progressive Distillation~\cite{salimans2022progressive} recursively compresses multi-step processes using ODE solvers. InstaFlow~\cite{liu2022flow,liu2023instaflow} progressively learns straighter flows, maintaining accuracy over larger distances with one-step predictions. Consistency Models~\cite{song2023consistency} unify distillation and ODE-based sampling, enabling 4–8 step generation via learned constraints. 
Moment Matching~\cite{salimans2025multistep} distill diffusion models by matching the conditional expectations of clean data given noisy data along the sampling trajectory.
The Shortcut model~\cite{frans2024one} accelerates inference via progressive self-distillation without requiring a pre-trained teacher. Inspired by this, our method integrates the Shortcut model with an autoregressive approach for efficient continuous-space modeling.

\section{Method}
In this section, we begin by introducing autoregressive image generation using a causal Transformer in discrete space (Section~\ref{sec:discrete}). We then revisit MAR~\cite{li2025mar}, which explores image generation in continuous space with autoregressive principles, and analyze its inference cost (Section~\ref{sec:MAR}). Finally, we present the proposed fast autoregressive model (FAR) and its variant, FAR-Causal, in Section~\ref{sec:far}.

\subsection{Generation with Discrete Representations}
\label{sec:discrete}

Building on the success of autoregressive language models like GPT~\cite{radford2019gpt2,brown2020gpt3}, recent image generation approaches~\cite{esser2021taming,yu2023language} have adopted similar methodologies. These methods first quantize an image into discrete tokens. Subsequently, a causal Transformer is employed to model the relationships between these tokens, enabling contextual image synthesis.

\noindent\textbf{Image Quantizer.} In general, an image quantizer~\cite{esser2021taming} consists of an encoder, a decoder, and a codebook. The encoder transforms an input image into a latent representation, which is then discretized by mapping it to the nearest entries in the codebook. The decoder reconstructs the image from the quantized tokens while minimizing information loss.

\noindent\textbf{Autoregressive Generation in Discrete Space.} As illustrated in Figure~\ref{fig:teaser-2}(b), a causal Transformer~\cite{radford2019gpt2,esser2021taming} predicts each token sequentially based on its preceding tokens. Formally, let $\boldsymbol{x}$ denote an image and $\{\boldsymbol{t}_1,...,\boldsymbol{t}_i,...,\boldsymbol{t}_N\}$ represent its flattened sequence of quantized tokens, where $\boldsymbol{t}_i \in \mathcal{C}$ corresponds to an entry in the codebook $\mathcal{C}$. The model learns a conditional probability distribution by minimizing the negative log-likelihood loss:
\begin{equation}
    \mathcal{L}=-\sum_{i=1}^N \log P_\theta\left(\boldsymbol{t}_i \mid \boldsymbol{t}_1, \boldsymbol{t}_2, \ldots, \boldsymbol{t}_{i-1}\right),
\end{equation}
where $\theta$ denotes the model parameters. This objective encourages the model to maximize the likelihood of correctly predicting each token given its preceding context.

\subsection{From Discrete to Continuous}
\label{sec:MAR}
Directly applying autoregressive generation from the language domain to the image domain typically necessitates image tokenizers that convert images into discrete tokens. While discrete representations enable efficient modeling with causal Transformers, the fixed-size codebook limits expressiveness, often resulting in information loss. Although modern image quantizers incorporate various optimization techniques—such as an exponential moving average strategy for refining the codebook~\cite{VQVAE2}, factorized code mechanisms for improved efficiency~\cite{VITVQGAN}, larger codebooks for enhanced expressiveness~\cite{zhu2024scaling}, and residual quantization for better reconstruction quality~\cite{RQVAE}—developing a highly effective image quantizer remains a significant challenge.

\noindent \textbf{Masked Autoregressive Model (MAR).} Recent work, MAR~\cite{li2025mar}, explores image generation within a continuous latent space defined by a VAE while preserving autoregressive principles—iteratively generating a set of tokens conditioned on previously generated tokens. As shown in Figure~\ref{fig:teaser-2}(a), MAR consists of an encoder, a decoder, and a diffusion head, with both the encoder and decoder consisting of a stack of self-attention blocks. 

\begin{figure*}[!t]
    \centering
\includegraphics[width=0.99\textwidth]{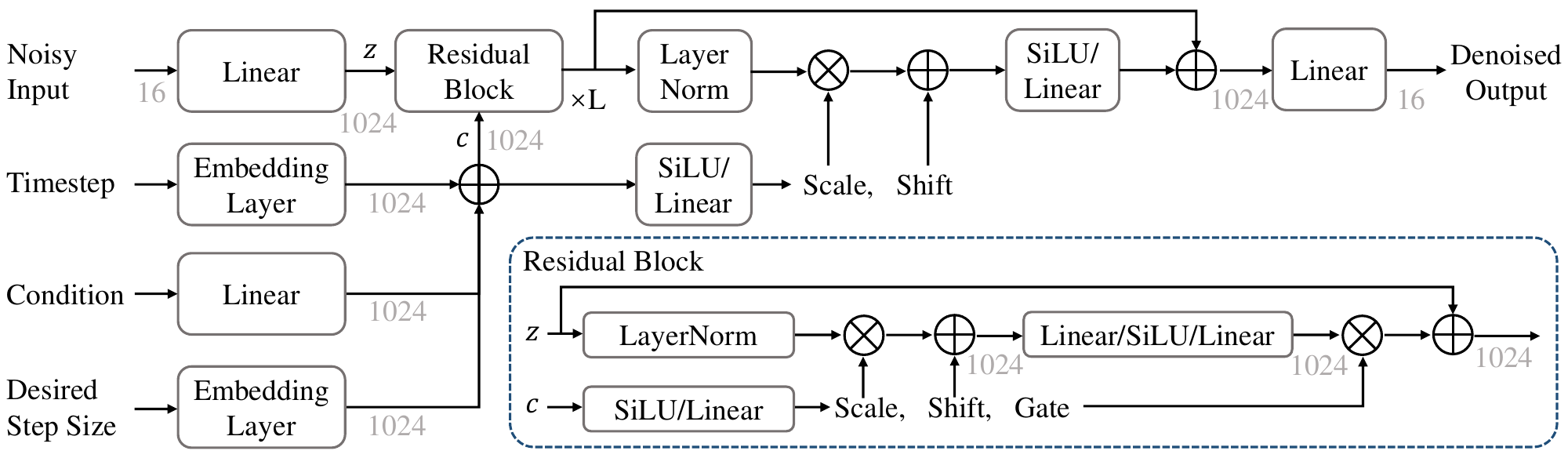}
    \caption{Architecture of the FAR head, a shortcut-based~\cite{frans2024one} network. The network processes a noisy token as input and produces a denoised output, guided jointly by a condition from the backbone, a desired step size, and a denoising timestep.} %
    \label{fig:head}
\end{figure*}

Specifically, MAR first employs a pre-trained VAE to encode an image into a latent representation, denoted as $\boldsymbol{Z}$\footnote{For clarity, we refer to the continuous elements in $\boldsymbol{Z}$ as ``image tokens''.}$\in \mathbb{R}^{h \times w \times d}$. Next, MAR randomly masks a subset of image tokens within $\boldsymbol{Z}$. Let $\mathcal{U} = \{\hat{\boldsymbol{z}}_{i}\}_{i=1}^{\hat{N}}$ and $\mathcal{M} = \{\boldsymbol{z}'_{i}\}_{i=1}^{N'}$ represent the sets of unmasked and masked tokens, respectively, where each $\hat{\boldsymbol{z}}_{i}$ and $\boldsymbol{z}'_{i}$ corresponds to a spatial element in $\boldsymbol{Z}$. Here, $\hat{N}$ denotes the number of unmasked tokens, while $N'$ represents the number of masked tokens. The encoder processes $\mathcal{U}$ to extract latent features. The decoder then takes as input both these latent features and a set of learnable tokens of size $N'$, each corresponding to a masked token in $\mathcal{M}$. It subsequently generates the features of these learnable tokens, referred to as conditions. Finally, the diffusion head, referred to as the MAR head in this work, independently performs a denoising process on each masked token $\boldsymbol{z}'_{i}$ using its corresponding condition.

\noindent \textbf{Inference Cost Analysis of MAR.} MAR follows an iterative inference process, requiring $K$ steps to complete ($K=64$ by default). In each step, it generates $M$\footnote{Notably, $M$ varies dynamically at different steps.} image tokens conditioned on the previously generated tokens. Specifically, the encoder extracts features from the existing generated tokens, the decoder processes these features along with the $M$ learnable tokens to produce conditions, and the diffusion head utilizes these conditions to produce $M$ predictions, each undergoing an $O$-step denoising process, where $O=100$ in the original implementation.

We use MAR-B~\cite{li2025mar} as an example to analyze the inference cost of MAR. In MAR-B, both the encoder and decoder comprise 12 attention blocks, while the diffusion head is a 6-layer MLP network. Figure~\ref{fig:teaser-1} presents a breakdown of the inference cost across different components, specifically comparing the encoder-decoder cost and the diffusion head cost when generating an image at a resolution of $(256, 256)$ (corresponding to $16 \times 16$ image tokens when using a VAE with a downsampling ratio of 16). 

\textit{Our analysis reveals that the diffusion head accounts for 63\% of the total inference cost.} This is primarily because generating each image token requires invoking the diffusion head $O$ times for an $O$-step denoising process. Given that $O=100$ in the original implementation, generating all $16 \times 16$ image tokens results in $100 \times 16 \times 16 = 25,600$ diffusion head calls, significantly contributing to the overall computation cost. To mitigate this, a more lightweight diffusion head is essential for reducing inference costs.

\subsection{Fast Autoregressive Model}
\label{sec:far}
We introduce the \underline{F}ast \underline{A}uto\underline{R}egressive model (FAR), designed to 1) optimize the inference efficiency of MAR; and 2) enable a causal Transformer for image generation, transitioning from discrete to continuous space. As illustrated in Figures~\ref{fig:teaser-2}(a) and (c), \textit{FAR} improves MAR by replacing the MAR head with the FAR head, allowing the generation of multiple tokens per iteration; additionally, it replaces the classification head in a causal Transformer with the FAR head, enabling single-token generation per iteration—a variant we refer to as \textit{FAR-Causal}. Both variants share the same head architecture and follow the same generation process—denoising one or multiple image tokens based on the given conditions, as detailed below.

\noindent\textbf{Condition Generation Process of FAR and FAR-Causal.} Given an image, a pre-trained VAE encodes it into a feature map $\boldsymbol{Z}\in\mathbb{R}^{h \times w \times d}$. For clarity, we refer to the continuous elements in $\boldsymbol{Z}$ as ``image tokens'', and denote the set of image tokens as $\{\boldsymbol{z}_i\}_{i=1}^N$ ($N = hw$), obtained by flattening $\boldsymbol{Z}$ in a raster order.

As shown in Figure~\ref{fig:teaser-2}(a), \textit{FAR} utilizes the encoder-decoder of MAR to extract conditions for the image tokens to be generated. As described in Section~\ref{sec:MAR}, MAR partitions the image token set $\{\boldsymbol{z}_i\}_{i=1}^N$ into an unmasked token set $\mathcal{U}$ of size $\hat{N}$, and a masked token set $\mathcal{M}$ of size $N'$. The encoder-decoder jointly models $\mathcal{U}$ and $\mathcal{M}$ to generate a set of conditions $\{\boldsymbol{c}_i\}_{i=1}^{N'}$, where each $\boldsymbol{c}_i$ corresponds to a masked token in $\mathcal{M}$. We formulate this process as:
\begin{equation}
\label{eq:MAR}
    \{\boldsymbol{c}_i\}_{i=1}^{N'} = \text{Decoder}(\text{Encoder}(\mathcal{U}), \mathcal{M}).
\end{equation}
Subsequently, each $\boldsymbol{c}_i$ is then independently processed by the FAR head. 

In contrast, as shown in Figure~\ref{fig:teaser-2}(c), \textit{FAR-Causal} extracts a condition $\boldsymbol{c}_i$ to predict $\boldsymbol{z}_i$, given its preceding image tokens $\{\boldsymbol{z}_{i-1},\boldsymbol{z}_{i-2},...,\boldsymbol{z}_{1}\}$, using a causal Transformer. This process is formulated as:
\begin{equation}
\label{eq:causal}
    \boldsymbol{c}_i = \text{CausalTransformer}(\boldsymbol{z}_{i-1},\boldsymbol{z}_{i-2},...,\boldsymbol{z}_{1}).
\end{equation}

For clarity, we omit the subscript $i$ and use $\boldsymbol{c}$ to represent the condition generated by Equation~\ref{eq:MAR} or Equation~\ref{eq:causal}, and $\boldsymbol{z}$ denotes the corresponding image token to predict.

\noindent\textbf{Generative Head Network.} 
Figure~\ref{fig:head} illustrates the structure of our head network. We present a novel lightweight shortcut head to enable few-step generation. Inspired by the shortcut model~\cite{frans2024one}, this head is conditioned on the condition $\boldsymbol{c}$, timestep $t$, and an additional desired step size $d$ during training, enabling the model to dynamically skip intermediate steps in the generation process. The optimization comprises two components: (1) a flow matching loss and (2) a consistency loss. The flow matching loss establishes an optimal transport path, guiding learning while ensuring that the denoising step remains flexible during inference. The loss $\mathcal{L}_{\text{FM}}$ and predicted target $\boldsymbol{v}$ are defined as follows:
\begin{align}
\mathcal{L}_{\text{FM}} &= \mathbb{E}_{\boldsymbol{z}_0, \boldsymbol{z}_1, t}\left[ 
 \left\lVert f_{\theta}\left( \boldsymbol{z}_t, \boldsymbol{c},0 \right) - \boldsymbol{v}\right\rVert_2^2\right], \\
 \boldsymbol{z}_t &= t \boldsymbol{z}_1 + \left(1-(1-\sigma_{\text{min}})t\right)\boldsymbol{z}_0, \\
 \boldsymbol{v} &= \boldsymbol{z}_1 - \left(1-\sigma_{\text{min}}\right)\boldsymbol{z}_0,
\end{align}
where $f_{\theta}$ represents the head network, $\boldsymbol{z}_0 \sim \mathcal{N}(\boldsymbol{0},\boldsymbol{I})$, $t\sim \mathcal{U}(0, 1)$ (i.e., uniformly sampled from $[0,1]$), $\boldsymbol{z}_1$ is the ground-truth token, and $\sigma_{\text{min}}$ is set to $1\times 10^{-5}$ to ensure consistency between training and sampling.

To enable few-step generation, we introduce an additional consistency loss $\mathcal{L}_{\text{Consist}}$, following~\cite{frans2024one}:
\begin{align}
    &\mathcal{L}_{\text{Consist}} = \mathbb{E}_{\boldsymbol{z}_0,\boldsymbol{z}_1,d,t} \left[ 
 \left\lVert f_{\theta}\left( \boldsymbol{z}_t, \boldsymbol{c},d \right) - \text{stopgrad}(\boldsymbol{v}_*)\right\rVert_2^2\right],
\end{align}
where $\boldsymbol{v}_*$ is calculated using the Exponential Moving Average (EMA) version of $f_{\theta}$, denoted as $f_{\text{EMA}}$. Specifically, 
\begin{equation}
    \begin{split}
        \boldsymbol{v}_{*} &= \frac{\boldsymbol{v}_t + \boldsymbol{v}_{t+\frac{d}{2}}}{2}, \\
        \boldsymbol{v}_t &= f_{{\text{EMA}}}(\boldsymbol{z}_t, \boldsymbol{c}, \frac{d}{2}), \\
        \boldsymbol{v}_{t+\frac{d}{2}} &= f_{\text{EMA}}(\tilde{\boldsymbol{z}}_{t+\frac{d}{2}}, \boldsymbol{c}, \frac{d}{2}),\\
        \tilde{\boldsymbol{z}}_{t+\frac{d}{2}}&=\boldsymbol{z}_t + \frac{d}{2}\boldsymbol{v}_t.
    \end{split}
\end{equation}
Unlike \cite{frans2024one}, which samples $d$ from a specific set, we uniformly sample it from $\mathcal{U}(0, 1)$, ensuring that $d = \min\{1 - t, \}$ to maintain valid trajectories.
The overall loss function combines both components:
\begin{align}
    \mathcal{L} = \mathcal{L}_{\text{FM}} + \mathcal{L}_{\text{Consist}}.
\end{align}

During inference, this method enables few-step generation, allowing the desired step size $d$ to be flexibly adjusted to balance speed and quality. Larger steps facilitate ultra-fast generation, while smaller steps enhance detail refinement—all without requiring retraining or architectural modifications. This adaptability makes the approach well-suited to varying computational budgets, outperforming both the traditional diffusion head~\cite{li2025mar} and existing distillation-based methods~\cite{salimans2022progressive} in efficiency and output quality. As a result, it is highly practical for real-time applications.

\section{Experiments}

\begin{table*}[!htp]
    \centering
    \begin{tabular}{l|ccccc}
    \toprule
          Method          & Type & Training Epochs & Parameters & FID$\downarrow$ & IS$\uparrow$ \\
    \midrule
      BigGAN-Deep~\cite{brock2018biggan}   &  GAN     & - & - & 6.95 & - \\
    \midrule
      ADM-G~\cite{dhariwal2021adm}         & Diffusion & 426 &  & 4.59 & 186.70 \\
      LDM-4-G~\cite{rombach2022ldm}         & Diffusion & - &  400M & 3.60 & 247.67 \\
      DiT-XL/2~\cite{peebles2023dit}         & Diffusion & 1400 & 675M & 2.27 & 278.24  \\
    \midrule
        JetFormer-B~\cite{tschannen2024jetformer}     &   Autoregressive    & 500 & 389M &  7.25 &    - \\
        JetFormer-L~\cite{tschannen2024jetformer}     &   Autoregressive    & 500 & 1.65B &  6.64 &   -  \\
     \midrule
    MaskGIT~\cite{chang2022maskgit}         & Maksed Autoregressive & 300 & 227M  & 4.02 & 355.60 \\
    MAR-B~\cite{li2025mar}         & Maksed Autoregressive & 800 & 172M  & 2.31 & 296.00 \\
        MAR-L~\cite{li2025mar}         & Maksed Autoregressive & 400 & 406M  & 1.98 &  290.30 \\
    \midrule
    FAR-B-Causal      &   Autoregressive    & 400 & 172M &  5.67 &    226.18 \\
    FAR-B      &   Masked Autoregressive    & 400 & 172M &  2.37 &   265.54 \\
    FAR-L      &   Masked Autoregressive    & 400 & 406M &  1.99  &   293.04 \\
    \bottomrule
    \end{tabular}
    \caption{Comparison of FAR, MAR~\cite{li2025mar}, and other generative models for $256 \times 256$ ImageNet generation, across various model scales. Notably, FAR-B (FID 2.37, 400 epochs) is 2.3$\times$ faster than its competitor MAR-B (FID 2.31, 800 epochs).
    }
    \label{tab:overall_compare}
\end{table*}

\subsection{Implementation Details}
Building on MAR's encoder-decoder design principles, we introduce FAR-B and FAR-L. FAR-B employs 12 attention blocks in both the encoder and decoder, while FAR-L uses 16 attention blocks. Both models share the same head network, a 6-layer MLP. For FAR-Causal, we adopt a 24-layer causal Transformer as the backbone while retaining the same head network. Additional details can be found in the supplementary material.

For a fair comparison with MAR~\cite{li2025mar}, we utilize its VAE to encode each image into a $16 \times 16 \times 16$ continuous latent representation. For optimization, we use the Adam optimizer~\cite{kingma2014adam,adamw} with a learning rate of $2 \times 10^{-4}$, which linearly warms up over the first 100 epochs. We apply a decoupled weight decay of 0.03 and set the momentum parameters $(\beta_1, \beta_2)$ to $(0.9, 0.999)$. Gradient norms are clipped at 1.0. Following standard practice, bias parameters and those in the FAR head are excluded from weight decay regularization. During training, we maintain an Exponential Moving Average (EMA) model with a decay rate of 0.9999. Since our primary goal is to demonstrate the effectiveness of the proposed method rather than achieving state-of-the-art results, we limit training to 400 epochs for most models. All models are trained on the ImageNet-1K dataset~\cite{deng2009imagenet} across $32\times$ NVIDIA V100 GPUs.

\noindent \textbf{Evaluation.}
To compare performance across different settings, we generate $50,000$ images using the EMA model and report the FID-50K score and Inception Score (IS). For FAR-B and FAR-L, we adopt sampling with a cosine masking schedule following MAR~\cite{li2025mar}. Additionally, we apply linear classifier-free guidance~\cite{ho2022classifier,li2025mar} during sampling.

\subsection{Main Results}
Table~\ref{tab:overall_compare} compares our model with state-of-the-art generative approaches for generating ImageNet images at a resolution of $256\times 256$. Our FAR-B achieves competitive performance (FID 2.37) compared to MAR-B (FID 2.31) while maintaining the same parameter count. Additionally, it offers a 2.3$\times$ speedup (Figure~\ref{fig:teaser-1}) and requires only 400 training epochs—half of MAR-B's 800 epochs. 

When scaling up to a larger encoder-decoder with 406M parameters, FAR-L achieves nearly identical performance (FID 1.99) to MAR-L (FID 1.98) while being 1.4$\times$ faster. Furthermore, FAR-L surpasses the larger and more extensively trained diffusion-based model, DiT-XL/2 (FID 2.27), which has 675M parameters and undergoes 1400 training epochs, by a FID score of 0.28.

Although our primary focus is FAR, which enhances the efficiency of MAR while maintaining nearly the same performance, we also introduce a variant, FAR-Causal, as detailed in Section~\ref{sec:far}. FAR-Causal consists of a causal Transformer and the proposed FAR head, transitioning from next-token prediction in discrete space to next-token distribution prediction in continuous space. Causal Transformers play a crucial role in language modeling—not only do they offer strong scalability and generalization capabilities, but they can also leverage KV cache to significantly accelerate inference. In Table~\ref{tab:overall_compare}, we report the performance of FAR-B-Causal, which achieves a respectable FID of 5.67 despite not using any tricks. Although its performance is lower than FAR-B, it retains key advantages of causal Transformers, such as efficient inference through KV cache utilization, and has the potential for improved performance by incorporating advanced optimization techniques from causal Transformers in language modeling.

\begin{figure*}[!t]
    \centering
\begin{tikzpicture}
  \begin{axis}[
    name=plot1,
    xlabel={Denoising Steps ($O$) Per Set Generation},
    ylabel={Head Network Cost Ratio (\%)},
    xmode=log,
    log basis x={2},
    grid=both,
    legend pos=north west,
    legend style={cells={anchor=west}},
    xtick={2,8,25,50,100},
    xticklabels={2,8,25,50,100},
    width=0.48\textwidth,
    height=0.3\textwidth,
    cycle list name=scicolors,
    yticklabel style={xshift=-0.1em},
    ylabel style={yshift=-0.8em},
    ymin=0, ymax=70
  ]
  
  \addplot[mark=o, thick, blue!50!black] coordinates {
    (100, {75.07018995285034/(75.07018995285034+43.819937229156494)*100})
    (50, {36.20159495544432/(36.20159495544432+44.10529687499999)*100})
    (25, {18.442789916992176/(18.442789916992176+44.347123657226554)*100})
    (8, {5.883636735916145/(5.883636735916145+44.16238081359866)*100})
    (2, {1.4981754860877996/(1.4981754860877996+45.2435046539307)*100})
  };
  \addlegendentry{$K=256$}
  
  \addplot[mark=square, thick, red!50!black] coordinates {
    (100, {19.626441741943367/(19.626441741943367+10.395763702392575)*100})
    (50, {9.812131835937507/(9.812131835937507+10.461802566528318)*100})
    (25, {5.248486518859863/(5.248486518859863+10.446907379150394)*100})
    (8,{1.6556666889190674/(1.6556666889190674+10.50822247314453)*100})
    (2, {0.40308019113540655/(0.40308019113540655+10.628186203002928)*100})
  };
  \addlegendentry{$K=64$}
  
  \addplot[mark=triangle, thick, green!50!black] coordinates {
    (100, {11.05268539428711/(11.05268539428711+5.159416809082033)*100})
    (50, {5.533219924926758/(5.533219924926758+5.1924848327636735)*100})
    (25, {2.8098826065063474/(2.8098826065063474+5.204744186401365)*100})
    (8,{0.9125898246765136/(0.9125898246765136+5.2369807434082025)*100})
    (2, {0.2451230721473694/(0.2451230721473694+5.238601730346679)*100})
  };
  \addlegendentry{$K=32$}
  
  \end{axis}
  
  \begin{axis}[
    name=plot2,
    at={($(plot1.east)+(1.5cm,0)$)},
    anchor=west,
    xlabel={Denoising Steps ($O$) Per Token Generation},
    ylabel={Head Network Cost Ratio (\%)},
    xmode=log,
    log basis x={2},
    grid=both,
    legend pos=north west,
    legend style={cells={anchor=west}},
    xtick={2,8,25,50,100},
    xticklabels={2,8,25,50,100},
    width=0.48\textwidth,
    height=0.3\textwidth,
    cycle list name=scicolors,
    yticklabel style={xshift=-0.1em},
    ylabel style={yshift=-0.8em},
    ymin=0, ymax=100
  ]
  
  \addplot[mark=o, thick, blue!50!black] coordinates {
    (100, {75.26778221130371/(75.26778221130371+53.80556797981262)*100})
    (50, {37.557328939437866/(37.557328939437866+54.743430852890015)*100})
    (25, {19.024914503097534/(19.024914503097534+54.99865245819092)*100})
    (8,{5.9914538860321045/(5.9914538860321045+56.8193922042846)*100})
    (2, {1.5979104042053223/(1.5979104042053223+54.73718237876892)*100})
  };
  \addlegendentry{w/o KV Cache}
  
  \addplot[mark=square, thick, red!50!black] coordinates {
    (100, {73.24612426757812/(73.24612426757812+6.128425121307373)*100})
    (50, {36.7383348941803/(36.7383348941803+6.122915983200073)*100})
    (25, {18.450525760650635/(18.450525760650635+6.118981122970581)*100})
    (8, {5.999629974365234/(5.999629974365234+ 6.1229047775268555)*100})
    (2, {1.557812213897705/(1.557812213897705+6.121169805526733)*100})
  };
  \addlegendentry{w/ KV Cache}
  
  \end{axis}
  
\end{tikzpicture}
\vspace{-3mm}
\caption{Analysis on the proportion of inference cost attributed to the head network in FAR (\textbf{left}) and FAR-Causal (\textbf{right}) under various settings. \textbf{Left}: Evaluation of three FAR variants, each using a different number of iterations ($K=32,64,256$) to generate an image. For each variant, we analyze the head network cost ratio by varying the number of denoising steps ($O=2,8,25,50,100$) required by the head network per image token generation.  \textbf{Right}: For each variant, we examine how the proportion of inference cost attributed to the head network changes as the number of denoising steps per image token generation varies.} %
    \label{fig:efficiency-analysis}
\end{figure*}
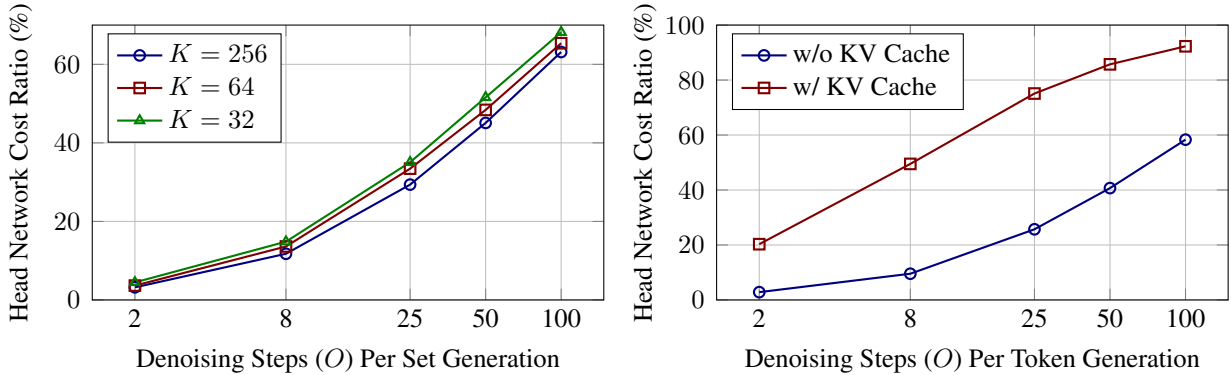

\subsection{Efficiency Analysis}
In Section~\ref{sec:MAR}, we conclude that the majority of MAR’s~\cite{li2025mar} inference cost stems from its diffusion-based head network—where generating a single image token requires calling the head network 100 times in its original implementation. This observation motivates the development of FAR, which leverages a more efficient shortcut head, reducing the number of head network calls per image token to just 8 times by default.

To further investigate efficiency, we analyze the proportion of inference cost attributed to the head network in both FAR and FAR-Causal, using FAR-B and FAR-B-Causal for evaluation. All measurements are conducted on a single NVIDIA A100 GPU with a batch size of 128.
\begin{itemize}
    \item \textit{FAR.} FAR follows MAR-style generation, where a set of tokens is generated at each iteration, conditioned on previously generated tokens. By default, FAR completes the process in $K=64$ iterations, with each token requiring $O=8$ head network calls per iteration. Figure~\ref{fig:efficiency-analysis} (left) examines the inference cost percentage of the head network by varying $K=32, 64, 256$ and $O=2,8,25,50,100$. The results indicate that as the number of denoising steps increases (i.e., a higher value of $O$ for denoising a single token), the head network progressively dominates the total inference time across all values of $K$. This evaluation confirms the key source of FAR’s inference speedup over MAR~\cite{li2025mar}—FAR calls the head network $O=8$ times per token generation, whereas MAR requires $O=100$ calls.
    \item \textit{FAR-Causal.} Unlike FAR, which generates a set of tokens at each iteration, FAR-Causal employs a causal Transformer that generates one token per iteration. Consequently, it requires a fixed 256 iterations to generate a latent representation of $16 \times 16$, corresponding to a $256 \times 256$ image. A key advantage of FAR-Causal is its ability to leverage the KV cache to accelerate inference. In Figure~\ref{fig:efficiency-analysis} (right), we analyze the proportion of inference cost attributed to the head network in FAR-Causal by varying the number of head network calls per token generation ($O=2,8,25,50,100$), comparing implementations with and without KV cache. The same trend is observed: as $O$ increases, the head network progressively dominates the total inference time. This highlights the importance of using a head network with fewer denoising steps to reduce overall inference cost.
\end{itemize}

\subsection{Ablation Studies}
Unless stated otherwise, we use FAR-B trained for 400 epochs to generate a $16 \times 16$ latent representation, corresponding to a $256 \times 256$ image, in all ablation studies.

\begin{table}[!t]
  \centering
  \begin{tabular}{cccc}
    \toprule
    AR Iterations & Denoising Steps & FID$\downarrow$ & IS$\uparrow$ \\
    \midrule
    \multirow{3}{*}{32} & 1 & 2.91& 273.18 \\
     & 4 & 2.70 & 279.66 \\
     & 8 & 2.69 & 278.78 \\
    \midrule
    \multirow{3}{*}{64} & 1 & 2.67& 261.86 \\
     & 4 & 2.56 & 284.66\\
     & 8 & 2.45 & 265.96 \\
    \midrule
    \multirow{3}{*}{256} & 1 & 2.55 & 259.46 \\
     & 4 & 2.40 & 264.74 \\
     & 8 & 2.37&  265.54\\
    \bottomrule
  \end{tabular}
  \vspace{-2mm}
  \caption{Ablation study on the impact of autoregressive (AR) iterations $K$ and denoising steps $O$ per token generation in FAR-B.}
  \vspace{-1mm}
  \label{tab:ar-iter-denoise-steps}
\end{table}

\noindent \textbf{The Impact of Autoregressive Iterations and Denoising Steps.} FAR follows MAR-style generation, where a set of image tokens is generated at each iteration, requiring $K$ iterations to complete the process. At each iteration, the FAR head performs an $O$-step denoising process to generate an image token. Table~\ref{tab:ar-iter-denoise-steps} examines the impact of different values of $K=32,64,256$ and $O=1,4,8$. Increasing $K$ from 32 to 256 gradually improves performance, reducing FID from 2.69 to 2.37 when $O=8$. However, we observe that $O=8$ provides only a marginal improvement over $O=4$. Thus, for applications requiring a balance between inference latency and performance, we recommend setting the denoising steps to 4.

\begin{table}[!t]
  \centering
  \begin{tabular}{ccc}
    \toprule
    Depth & FID$\downarrow$ & IS$\uparrow$ \\
    \midrule
    3 & {3.27} & {269.93} \\
    6 & {2.69} & {278.78} \\
    \bottomrule
  \end{tabular}
  \vspace{-2mm}
  \caption{Ablation study on FAR head depth.}
  \vspace{-3mm}
  \label{fig:diff-head-depths}
\end{table}

\noindent \textbf{Depth of FAR Head.} Table~\ref{fig:diff-head-depths} examines the impact of FAR head depth. We evaluate two variants: a 6-layer head and a 3-layer head. This ablation study is conducted on FAR-B with $K=32$ iterations and $O=8$ denoising steps. By default, we use the 6-layer head.

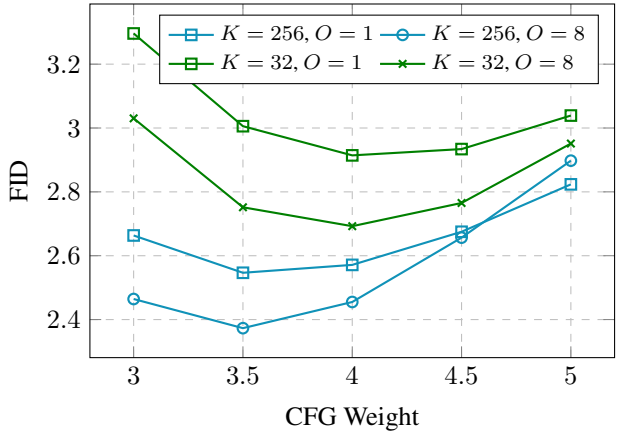
\begin{figure}[!t]
\centering
\begin{tikzpicture}
\begin{axis}[
    width=0.49\textwidth,
    height=0.36\textwidth,
    xlabel={CFG Weight},
    ylabel={FID},
    grid=both,
    legend pos=north east,
    legend style={font=\footnotesize, legend columns=2, cells={anchor=west}},
    xmin=2.8, xmax=5.2,
    xtick={3.0, 3.5, 4.0, 4.5, 5.0},
    ytick={2.2, 2.4, 2.6, 2.8, 3.0, 3.2, 3.4},
    ylabel style={yshift=-0.8em},
    ymajorgrids=true,
    grid style=dashed,
    fill=teal!70,
]

\addplot[color=cyan!70!black, mark=square, thick] coordinates {
    (3.0, 2.6634) (3.5, 2.5466) (4.0, 2.5713) (4.5, 2.6750) (5.0, 2.8235)
};
\addlegendentry{$K=256$, $O=1$}

\addplot[color=cyan!70!black, mark=o, thick] coordinates {
    (3.0, 2.4644) (3.5, 2.3731) (4.0, 2.4551) (4.5, 2.6566) (5.0, 2.8977)
};
\addlegendentry{$K=256$, $O=8$}

\addplot[color=green!50!black, mark=square, thick] coordinates {
    (3.0, 3.2961) (3.5, 3.0057) (4.0, 2.9143) (4.5, 2.9340) (5.0, 3.0390)
};
\addlegendentry{$K=32$, $O=1$}

\addplot[color=green!50!black, mark=x, thick] coordinates {
    (3.0, 3.0303) (3.5, 2.7517) (4.0, 2.6925) (4.5, 2.7652) (5.0, 2.9513)
};
\addlegendentry{$K=32$, $O=8$}

\end{axis}
\end{tikzpicture}
\vspace{-3mm}
\caption{The impact of different CFG weights on FID for FAR-B under various configurations of autoregressive iterations ($K=32, 256$) and denoising steps ($O=1, 8$) per token generation. }
\vspace{-3mm}
\label{fig:fid_comparison}
\end{figure}

\noindent \textbf{Classifier-Free Guidance (CFG).} 
Figure~\ref{fig:fid_comparison} illustrates the impact of different CFG weights on the performance of FAR-B under various configurations of autoregressive iterations ($K=32, 256$) and denoising steps ($O=1, 8$) per token generation.

\begin{figure*}[h]
    \centering
    \includegraphics[width=\textwidth]{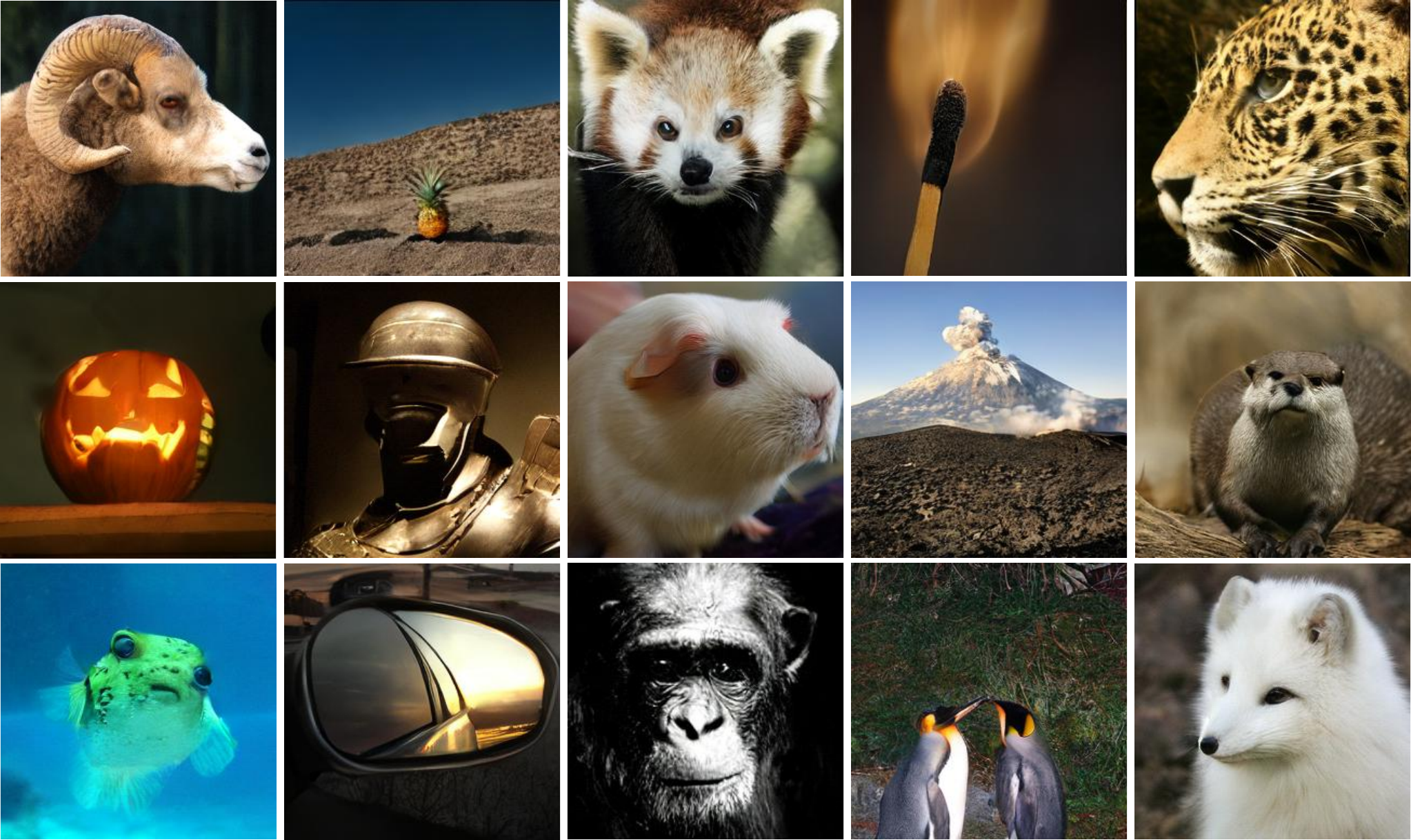}
    \caption{Visual examples of images generated by FAR-B.} %
    \label{fig:visual-example}
\end{figure*}

\noindent\textbf{Shortcut vs. Flow Matching.}
In Section~\ref{sec:far}, we introduce an efficient shortcut head that performs an 8-step denoising process to generate an image token. Here, we explore a head variant that leverages flow matching~\cite{lipman2022flow,liu2022flow} for denoising noisy tokens. All other configurations, including the head network architecture, training epochs, remain unchanged compared to our default head implementation. Table~\ref{tab:abl-two-stage-one-stage} compares FAR-B with a shortcut head to FAR-B with a flow matching head across various denoising steps ($O=1,2,4,8$) per token generation. For computational efficiency, both head variants utilize a 3-layer MLP network in this study. The results in Table~\ref{tab:abl-two-stage-one-stage} indicate that employing the FAR head with shortcut significantly outperforms its counterpart with flow matching across various denoising steps, particularly in extremely few-step denoising scenarios where $O=1$ and $O=2$.

Since the network architectures of the flow matching head and the shortcut head are nearly identical—differing only by an additional embedding layer in the shortcut head to encode the ``desired step size'', as shown in Figure~\ref{fig:head}. Here, we explore an alternative approach, termed ``flow-matching$\rightarrow$shortcut''. In this setup, we first pre-train FAR-B with a flow matching head for 400 epochs, then adaptively modify the head network structure to a shortcut head, and finally fine-tune the model for 10 epochs. As shown in Table~\ref{tab:abl-two-stage-one-stage}, this training paradigm improves the performance of the flow matching head but still lags behind the shortcut head.

\begin{table}[!t]
  \centering
  \small
  \begin{tabular}{ccccccc}
    \toprule
    \multirow{2.5}{*}{Steps} & \multicolumn{2}{c}{Flow Matching} & \multicolumn{2}{c}{Shortcut} & \multicolumn{2}{c}{FM$\rightarrow$SC} \\
    \cmidrule(lr){2-3} \cmidrule(lr){4-5} \cmidrule(lr){6-7}
     & FID$\downarrow$ & IS$\uparrow$ & FID$\downarrow$ & IS$\uparrow$ & FID$\downarrow$ & IS$\uparrow$ \\
    \midrule
    1 & 113.60 & 10.19 & 3.86 & 282.77 & 21.79 & 114.91 \\
    2 & 46.48 & 65.22 & 3.46 & 299.94 & 4.46 & 233.60 \\
    4 & 14.19 & 210.08 & 3.38 & 303.47 & 3.36 & 226.20 \\
    8 & 5.24 & 270.44 & 3.27 & 269.94 & 3.43 & 234.51 \\
    \bottomrule
  \end{tabular}
  \vspace{-1mm}
  \caption{Performance comparison between flow matching-based FAR-B and shortcut-based FAR-B, both trained for 400 epochs. ``FM$\rightarrow$SC'' denotes a two-stage approach where FAR-B is pre-trained with flow matching for 400 epochs, followed by fine-tuning with shortcut for 10 epochs.}
  \label{tab:abl-two-stage-one-stage}
\end{table}

\noindent\textbf{Visualizations.} Figure~\ref{fig:visual-example} provides visual examples of images generated by FAR-B.

\section{Conclusion}
This paper begins by analyzing the inference cost of the Masked AutoRegressive (MAR) model, an autoregressive framework operating in a continuous space, and identifies its diffusion head as the primary computational bottleneck, contributing to the majority of inference cost due to its repeated invocation for generating each image token. Motivated by this, we propose Fast AutoRegressive (FAR), a novel autoregressive framework for continuous-space image generation that significantly improves the inference efficiency of MAR while maintaining high generation quality. This efficiency gain stems from an efficient shortcut-based head, which enables faster sampling with fewer denoising steps per token generation. Experiments on ImageNet show that FAR achieves 2.3$\times$ faster inference than MAR while maintaining competitive FID and IS scores. Additionally, FAR seamlessly integrates with causal Transformers, allowing them to operate in a continuous space without architectural modifications. By addressing both efficiency and complexity challenges, FAR establishes a practical paradigm for high-fidelity image synthesis, advancing autoregressive modeling in visual generation.

{
    \small
    \bibliographystyle{ieeenat_fullname}
    \bibliography{main}

\begin{thebibliography}{62}
\providecommand{\natexlab}[1]{#1}
\providecommand{\url}[1]{\texttt{#1}}
\expandafter\ifx\csname urlstyle\endcsname\relax
  \providecommand{\doi}[1]{doi: #1}\else
  \providecommand{\doi}{doi: \begingroup \urlstyle{rm}\Url}\fi

\bibitem[Abdin et~al.(2024)Abdin, Jacobs, Awan, Aneja, Awadallah, Awadalla,
  Bach, Bahree, Bakhtiari, Behl, et~al.]{abdin2024phi}
Marah Abdin, Sam~Ade Jacobs, Ammar~Ahmad Awan, Jyoti Aneja, Ahmed Awadallah,
  Hany Awadalla, Nguyen Bach, Amit Bahree, Arash Bakhtiari, Harkirat Behl,
  et~al.
\newblock Phi-3 technical report: A highly capable language model locally on
  your phone.
\newblock \emph{arXiv preprint arXiv:2404.14219}, 2024.

\bibitem[Achiam et~al.(2023)Achiam, Adler, Agarwal, Ahmad, Akkaya, Aleman,
  Almeida, Altenschmidt, Altman, Anadkat, et~al.]{achiam2023gpt}
Josh Achiam, Steven Adler, Sandhini Agarwal, Lama Ahmad, Ilge Akkaya,
  Florencia~Leoni Aleman, Diogo Almeida, Janko Altenschmidt, Sam Altman,
  Shyamal Anadkat, et~al.
\newblock Gpt-4 technical report.
\newblock \emph{arXiv preprint arXiv:2303.08774}, 2023.

\bibitem[Anil et~al.(2023)Anil, Dai, Firat, Johnson, Lepikhin, Passos, Shakeri,
  Taropa, Bailey, Chen, et~al.]{anil2023palm}
Rohan Anil, Andrew~M Dai, Orhan Firat, Melvin Johnson, Dmitry Lepikhin,
  Alexandre Passos, Siamak Shakeri, Emanuel Taropa, Paige Bailey, Zhifeng Chen,
  et~al.
\newblock Palm 2 technical report.
\newblock \emph{arXiv preprint arXiv:2305.10403}, 2023.

\bibitem[Bai et~al.(2023)Bai, Bai, Chu, Cui, Dang, Deng, Fan, Ge, Han, Huang,
  et~al.]{bai2023qwen}
Jinze Bai, Shuai Bai, Yunfei Chu, Zeyu Cui, Kai Dang, Xiaodong Deng, Yang Fan,
  Wenbin Ge, Yu Han, Fei Huang, et~al.
\newblock Qwen technical report.
\newblock \emph{arXiv preprint arXiv:2309.16609}, 2023.

\bibitem[Bao et~al.(2022{\natexlab{a}})Bao, Li, Sun, Zhu, and
  Zhang]{bao2022estimating}
Fan Bao, Chongxuan Li, Jiacheng Sun, Jun Zhu, and Bo Zhang.
\newblock Estimating the optimal covariance with imperfect mean in diffusion
  probabilistic models.
\newblock \emph{ICML}, 2022{\natexlab{a}}.

\bibitem[Bao et~al.(2022{\natexlab{b}})Bao, Li, Zhu, and
  Zhang]{bao2022analytic}
Fan Bao, Chongxuan Li, Jun Zhu, and Bo Zhang.
\newblock Analytic-dpm: an analytic estimate of the optimal reverse variance in
  diffusion probabilistic models.
\newblock \emph{arXiv preprint arXiv:2201.06503}, 2022{\natexlab{b}}.

\bibitem[Berthelot et~al.(2023)Berthelot, Autef, Lin, Yap, Zhai, Hu, Zheng,
  Talbot, and Gu]{berthelot2023tract}
David Berthelot, Arnaud Autef, Jierui Lin, Dian~Ang Yap, Shuangfei Zhai, Siyuan
  Hu, Daniel Zheng, Walter Talbot, and Eric Gu.
\newblock Tract: Denoising diffusion models with transitive closure
  time-distillation.
\newblock \emph{arXiv preprint arXiv:2303.04248}, 2023.

\bibitem[Brock et~al.(2018)Brock, Donahue, and Simonyan]{brock2018biggan}
Andrew Brock, Jeff Donahue, and Karen Simonyan.
\newblock Large scale gan training for high fidelity natural image synthesis.
\newblock \emph{arXiv preprint arXiv:1809.11096}, 2018.

\bibitem[Brown et~al.(2020{\natexlab{a}})Brown, Mann, Ryder, Subbiah, Kaplan,
  Dhariwal, Neelakantan, Shyam, Sastry, Askell, et~al.]{brown2020gpt3}
Tom Brown, Benjamin Mann, Nick Ryder, Melanie Subbiah, Jared~D Kaplan, Prafulla
  Dhariwal, Arvind Neelakantan, Pranav Shyam, Girish Sastry, Amanda Askell,
  et~al.
\newblock Language models are few-shot learners.
\newblock \emph{Advances in neural information processing systems},
  33:\penalty0 1877--1901, 2020{\natexlab{a}}.

\bibitem[Brown et~al.(2020{\natexlab{b}})Brown, Mann, Ryder, Subbiah, Kaplan,
  Dhariwal, Neelakantan, Shyam, Sastry, Askell, et~al.]{brown2020language}
Tom Brown, Benjamin Mann, Nick Ryder, Melanie Subbiah, Jared~D Kaplan, Prafulla
  Dhariwal, Arvind Neelakantan, Pranav Shyam, Girish Sastry, Amanda Askell,
  et~al.
\newblock Language models are few-shot learners.
\newblock \emph{Advances in neural information processing systems},
  33:\penalty0 1877--1901, 2020{\natexlab{b}}.

\bibitem[Chang et~al.(2022)Chang, Zhang, Jiang, Liu, and
  Freeman]{chang2022maskgit}
Huiwen Chang, Han Zhang, Lu Jiang, Ce Liu, and William~T Freeman.
\newblock Maskgit: Masked generative image transformer.
\newblock In \emph{Proceedings of the IEEE/CVF conference on computer vision
  and pattern recognition}, pages 11315--11325, 2022.

\bibitem[Chen et~al.(2020)Chen, Radford, Child, Wu, Jun, Luan, and
  Sutskever]{chen2020generative}
Mark Chen, Alec Radford, Rewon Child, Jeffrey Wu, Heewoo Jun, David Luan, and
  Ilya Sutskever.
\newblock Generative pretraining from pixels.
\newblock In \emph{International conference on machine learning}, pages
  1691--1703. PMLR, 2020.

\bibitem[Chowdhery et~al.(2023)Chowdhery, Narang, Devlin, Bosma, Mishra,
  Roberts, Barham, Chung, Sutton, Gehrmann, et~al.]{chowdhery2023palm}
Aakanksha Chowdhery, Sharan Narang, Jacob Devlin, Maarten Bosma, Gaurav Mishra,
  Adam Roberts, Paul Barham, Hyung~Won Chung, Charles Sutton, Sebastian
  Gehrmann, et~al.
\newblock Palm: Scaling language modeling with pathways.
\newblock \emph{Journal of Machine Learning Research}, 24\penalty0
  (240):\penalty0 1--113, 2023.

\bibitem[Deng et~al.(2009)Deng, Dong, Socher, Li, Li, and
  Fei-Fei]{deng2009imagenet}
Jia Deng, Wei Dong, Richard Socher, Li-Jia Li, Kai Li, and Li Fei-Fei.
\newblock Imagenet: A large-scale hierarchical image database.
\newblock In \emph{CVPR}, 2009.

\bibitem[Dhariwal and Nichol(2021)]{dhariwal2021adm}
Prafulla Dhariwal and Alexander Nichol.
\newblock Diffusion models beat gans on image synthesis.
\newblock \emph{Advances in neural information processing systems},
  34:\penalty0 8780--8794, 2021.

\bibitem[Dong et~al.(2023)Dong, Han, Peng, Qi, Ge, Yang, Zhao, Sun, Zhou, Wei,
  et~al.]{dong2023dreamllm}
Runpei Dong, Chunrui Han, Yuang Peng, Zekun Qi, Zheng Ge, Jinrong Yang, Liang
  Zhao, Jianjian Sun, Hongyu Zhou, Haoran Wei, et~al.
\newblock Dreamllm: Synergistic multimodal comprehension and creation.
\newblock \emph{arXiv preprint arXiv:2309.11499}, 2023.

\bibitem[Dubey et~al.(2024)Dubey, Jauhri, Pandey, Kadian, Al-Dahle, Letman,
  Mathur, Schelten, Yang, Fan, et~al.]{dubey2024llama}
Abhimanyu Dubey, Abhinav Jauhri, Abhinav Pandey, Abhishek Kadian, Ahmad
  Al-Dahle, Aiesha Letman, Akhil Mathur, Alan Schelten, Amy Yang, Angela Fan,
  et~al.
\newblock The llama 3 herd of models.
\newblock \emph{arXiv preprint arXiv:2407.21783}, 2024.

\bibitem[Esser et~al.(2021)Esser, Rombach, and Ommer]{esser2021taming}
Patrick Esser, Robin Rombach, and Bjorn Ommer.
\newblock Taming transformers for high-resolution image synthesis.
\newblock In \emph{Proceedings of the IEEE/CVF conference on computer vision
  and pattern recognition}, pages 12873--12883, 2021.

\bibitem[Frans et~al.(2024)Frans, Hafner, Levine, and Abbeel]{frans2024one}
Kevin Frans, Danijar Hafner, Sergey Levine, and Pieter Abbeel.
\newblock One step diffusion via shortcut models.
\newblock \emph{arXiv preprint arXiv:2410.12557}, 2024.

\bibitem[Ge et~al.(2024)Ge, Zhao, Zhu, Ge, Yi, Song, Li, Ding, and
  Shan]{ge2024seed}
Yuying Ge, Sijie Zhao, Jinguo Zhu, Yixiao Ge, Kun Yi, Lin Song, Chen Li,
  Xiaohan Ding, and Ying Shan.
\newblock Seed-x: Multimodal models with unified multi-granularity
  comprehension and generation.
\newblock \emph{arXiv preprint arXiv:2404.14396}, 2024.

\bibitem[Gu et~al.(2023)Gu, Zhai, Zhang, Liu, and Susskind]{gu2023boot}
Jiatao Gu, Shuangfei Zhai, Yizhe Zhang, Lingjie Liu, and Joshua~M Susskind.
\newblock Boot: Data-free distillation of denoising diffusion models with
  bootstrapping.
\newblock In \emph{ICML 2023 Workshop on Structured Probabilistic Inference \&
  Generative Modeling}, 2023.

\bibitem[Ho and Salimans(2022)]{ho2022classifier}
Jonathan Ho and Tim Salimans.
\newblock Classifier-free diffusion guidance.
\newblock In \emph{arXiv preprint arXiv:2207.12598}, 2022.

\bibitem[Karras et~al.(2022)Karras, Aittala, Aila, and
  Laine]{karras2022elucidating}
Tero Karras, Miika Aittala, Timo Aila, and Samuli Laine.
\newblock Elucidating the design space of diffusion-based generative models.
\newblock In \emph{NeurIPS}, 2022.

\bibitem[Kingma and Ba(2014)]{kingma2014adam}
Diederik~P Kingma and Jimmy Ba.
\newblock Adam: A method for stochastic optimization.
\newblock \emph{arXiv preprint arXiv:1412.6980}, 2014.

\bibitem[Kingma and Welling(2014)]{kingma2013auto}
Diederik~P Kingma and Max Welling.
\newblock Auto-encoding variational bayes.
\newblock In \emph{ICLR}, 2014.

\bibitem[Lee et~al.(2022{\natexlab{a}})Lee, Kim, Kim, Cho, and Han]{RQVAE}
Doyup Lee, Chiheon Kim, Saehoon Kim, Minsu Cho, and Wook-Shin Han.
\newblock Autoregressive image generation using residual quantization.
\newblock In \emph{Proceedings of the IEEE/CVF Conference on Computer Vision
  and Pattern Recognition}, pages 11523--11532, 2022{\natexlab{a}}.

\bibitem[Lee et~al.(2022{\natexlab{b}})Lee, Kim, Kim, Cho, and
  Han]{lee2022autoregressive}
Doyup Lee, Chiheon Kim, Saehoon Kim, Minsu Cho, and Wook-Shin Han.
\newblock Autoregressive image generation using residual quantization.
\newblock In \emph{Proceedings of the IEEE/CVF Conference on Computer Vision
  and Pattern Recognition}, pages 11523--11532, 2022{\natexlab{b}}.

\bibitem[Li et~al.(2025)Li, Tian, Li, Deng, and He]{li2025mar}
Tianhong Li, Yonglong Tian, He Li, Mingyang Deng, and Kaiming He.
\newblock Autoregressive image generation without vector quantization.
\newblock \emph{Advances in Neural Information Processing Systems},
  37:\penalty0 56424--56445, 2025.

\bibitem[Lipman et~al.(2022)Lipman, Chen, Ben-Hamu, Nickel, and
  Le]{lipman2022flow}
Yaron Lipman, Ricky~TQ Chen, Heli Ben-Hamu, Maximilian Nickel, and Matt Le.
\newblock Flow matching for generative modeling.
\newblock \emph{arXiv preprint arXiv:2210.02747}, 2022.

\bibitem[Liu et~al.(2022)Liu, Ren, Lin, and Zhao]{liu2022pseudo}
Luping Liu, Yi Ren, Zhijie Lin, and Zhou Zhao.
\newblock Pseudo numerical methods for diffusion models on manifolds.
\newblock In \emph{ICLR}, 2022.

\bibitem[Liu et~al.(2023{\natexlab{a}})Liu, Gong, and Liu]{liu2022flow}
Xingchao Liu, Chengyue Gong, and Qiang Liu.
\newblock Flow straight and fast: Learning to generate and transfer data with
  rectified flow.
\newblock In \emph{ICLR}, 2023{\natexlab{a}}.

\bibitem[Liu et~al.(2023{\natexlab{b}})Liu, Zhang, Ma, Peng, and
  Liu]{liu2023instaflow}
Xingchao Liu, Xiwen Zhang, Jianzhu Ma, Jian Peng, and Qiang Liu.
\newblock Instaflow: One step is enough for high-quality diffusion-based
  text-to-image generation.
\newblock \emph{arXiv preprint arXiv:2309.06380}, 2023{\natexlab{b}}.

\bibitem[Loshchilov and Hutter(2017)]{adamw}
Ilya Loshchilov and Frank Hutter.
\newblock Decoupled weight decay regularization.
\newblock \emph{arXiv preprint arXiv:1711.05101}, 2017.

\bibitem[Lu et~al.(2022{\natexlab{a}})Lu, Zhou, Bao, Chen, Li, and
  Zhu]{lu2022dpm}
Cheng Lu, Yuhao Zhou, Fan Bao, Jianfei Chen, Chongxuan Li, and Jun Zhu.
\newblock Dpm-solver: A fast ode solver for diffusion probabilistic model
  sampling in around 10 steps.
\newblock \emph{Advances in Neural Information Processing Systems},
  35:\penalty0 5775--5787, 2022{\natexlab{a}}.

\bibitem[Lu et~al.(2022{\natexlab{b}})Lu, Zhou, Bao, Chen, Li, and
  Zhu]{lu2022dpm++}
Cheng Lu, Yuhao Zhou, Fan Bao, Jianfei Chen, Chongxuan Li, and Jun Zhu.
\newblock Dpm-solver++: Fast solver for guided sampling of diffusion
  probabilistic models.
\newblock \emph{arXiv preprint arXiv:2211.01095}, 2022{\natexlab{b}}.

\bibitem[Luhman and Luhman(2021)]{luhman2021knowledge}
Eric Luhman and Troy Luhman.
\newblock Knowledge distillation in iterative generative models for improved
  sampling speed.
\newblock \emph{arXiv preprint arXiv:2101.02388}, 2021.

\bibitem[Meng et~al.(2023)Meng, Rombach, Gao, Kingma, Ermon, Ho, and
  Salimans]{meng2023distillation}
Chenlin Meng, Robin Rombach, Ruiqi Gao, Diederik Kingma, Stefano Ermon,
  Jonathan Ho, and Tim Salimans.
\newblock On distillation of guided diffusion models.
\newblock In \emph{CVPR}, 2023.

\bibitem[Peebles and Xie(2023)]{peebles2023dit}
William Peebles and Saining Xie.
\newblock Scalable diffusion models with transformers.
\newblock In \emph{Proceedings of the IEEE/CVF international conference on
  computer vision}, pages 4195--4205, 2023.

\bibitem[Radford et~al.(2018)Radford, Narasimhan, Salimans, and
  Sutskever]{radford2018improving}
Alec Radford, Karthik Narasimhan, Tim Salimans, and Ilya Sutskever.
\newblock Improving language understanding by generative pre-training.
\newblock \emph{OpenAI}, 2018.
\newblock
  \url{https://cdn.openai.com/research-covers/language-unsupervised/language_understanding_paper.pdf}.

\bibitem[Radford et~al.(2019)Radford, Wu, Child, Luan, Amodei, Sutskever,
  et~al.]{radford2019gpt2}
Alec Radford, Jeffrey Wu, Rewon Child, David Luan, Dario Amodei, Ilya
  Sutskever, et~al.
\newblock Language models are unsupervised multitask learners.
\newblock \emph{OpenAI blog}, 1\penalty0 (8):\penalty0 9, 2019.

\bibitem[Razavi et~al.(2019)Razavi, Van~den Oord, and Vinyals]{VQVAE2}
Ali Razavi, Aaron Van~den Oord, and Oriol Vinyals.
\newblock Generating diverse high-fidelity images with vq-vae-2.
\newblock \emph{Advances in neural information processing systems}, 32, 2019.

\bibitem[Rombach et~al.(2022)Rombach, Blattmann, Lorenz, Esser, and
  Ommer]{rombach2022ldm}
Robin Rombach, Andreas Blattmann, Dominik Lorenz, Patrick Esser, and Bj{\"o}rn
  Ommer.
\newblock High-resolution image synthesis with latent diffusion models.
\newblock In \emph{Proceedings of the IEEE/CVF conference on computer vision
  and pattern recognition}, pages 10684--10695, 2022.

\bibitem[Salimans and Ho(2022)]{salimans2022progressive}
Tim Salimans and Jonathan Ho.
\newblock Progressive distillation for fast sampling of diffusion models.
\newblock In \emph{ICLR}, 2022.

\bibitem[Salimans et~al.(2025)Salimans, Mensink, Heek, and
  Hoogeboom]{salimans2025multistep}
Tim Salimans, Thomas Mensink, Jonathan Heek, and Emiel Hoogeboom.
\newblock Multistep distillation of diffusion models via moment matching.
\newblock \emph{Advances in Neural Information Processing Systems},
  37:\penalty0 36046--36070, 2025.

\bibitem[Sohn et~al.(2015)Sohn, Lee, and Yan]{sohn2015learning}
Kihyuk Sohn, Honglak Lee, and Xinchen Yan.
\newblock Learning structured output representation using deep conditional
  generative models.
\newblock \emph{Advances in neural information processing systems}, 28, 2015.

\bibitem[Song et~al.(2023)Song, Dhariwal, Chen, and
  Sutskever]{song2023consistency}
Yang Song, Prafulla Dhariwal, Mark Chen, and Ilya Sutskever.
\newblock Consistency models.
\newblock In \emph{ICML}, 2023.

\bibitem[Tang et~al.(2024)Tang, Wu, Yang, Xie, Chen, Chen, Zhang, Cai, Lu, and
  Han]{tang2024hart}
Haotian Tang, Yecheng Wu, Shang Yang, Enze Xie, Junsong Chen, Junyu Chen,
  Zhuoyang Zhang, Han Cai, Yao Lu, and Song Han.
\newblock Hart: Efficient visual generation with hybrid autoregressive
  transformer.
\newblock \emph{arXiv preprint arXiv:2410.10812}, 2024.

\bibitem[Team et~al.(2023)Team, Anil, Borgeaud, Wu, Alayrac, Yu, Soricut,
  Schalkwyk, Dai, Hauth, et~al.]{team2023gemini}
Gemini Team, Rohan Anil, Sebastian Borgeaud, Yonghui Wu, Jean-Baptiste Alayrac,
  Jiahui Yu, Radu Soricut, Johan Schalkwyk, Andrew~M Dai, Anja Hauth, et~al.
\newblock Gemini: a family of highly capable multimodal models.
\newblock \emph{arXiv preprint arXiv:2312.11805}, 2023.

\bibitem[Touvron et~al.(2023{\natexlab{a}})Touvron, Lavril, Izacard, Martinet,
  Lachaux, Lacroix, Rozi{\`e}re, Goyal, Hambro, Azhar,
  et~al.]{touvron2023llama}
Hugo Touvron, Thibaut Lavril, Gautier Izacard, Xavier Martinet, Marie-Anne
  Lachaux, Timoth{\'e}e Lacroix, Baptiste Rozi{\`e}re, Naman Goyal, Eric
  Hambro, Faisal Azhar, et~al.
\newblock Llama: Open and efficient foundation language models.
\newblock \emph{arXiv preprint arXiv:2302.13971}, 2023{\natexlab{a}}.

\bibitem[Touvron et~al.(2023{\natexlab{b}})Touvron, Martin, Stone, Albert,
  Almahairi, Babaei, Bashlykov, Batra, Bhargava, Bhosale,
  et~al.]{touvron2023llama2}
Hugo Touvron, Louis Martin, Kevin Stone, Peter Albert, Amjad Almahairi, Yasmine
  Babaei, Nikolay Bashlykov, Soumya Batra, Prajjwal Bhargava, Shruti Bhosale,
  et~al.
\newblock Llama 2: Open foundation and fine-tuned chat models.
\newblock \emph{arXiv preprint arXiv:2307.09288}, 2023{\natexlab{b}}.

\bibitem[Tschannen et~al.(2024{\natexlab{a}})Tschannen, Eastwood, and
  Mentzer]{tschannen2024givt}
Michael Tschannen, Cian Eastwood, and Fabian Mentzer.
\newblock Givt: Generative infinite-vocabulary transformers.
\newblock In \emph{European Conference on Computer Vision}, pages 292--309.
  Springer, 2024{\natexlab{a}}.

\bibitem[Tschannen et~al.(2024{\natexlab{b}})Tschannen, Pinto, and
  Kolesnikov]{tschannen2024jetformer}
Michael Tschannen, Andr{\'e}~Susano Pinto, and Alexander Kolesnikov.
\newblock Jetformer: An autoregressive generative model of raw images and text.
\newblock \emph{arXiv preprint arXiv:2411.19722}, 2024{\natexlab{b}}.

\bibitem[Van Den~Oord et~al.(2017)Van Den~Oord, Vinyals, et~al.]{van2017neural}
Aaron Van Den~Oord, Oriol Vinyals, et~al.
\newblock Neural discrete representation learning.
\newblock \emph{Advances in neural information processing systems}, 30, 2017.

\bibitem[Xiao et~al.(2022)Xiao, Kreis, and Vahdat]{xiao2021tackling}
Zhisheng Xiao, Karsten Kreis, and Arash Vahdat.
\newblock Tackling the generative learning trilemma with denoising diffusion
  gans.
\newblock In \emph{ICLR}, 2022.

\bibitem[Yang et~al.(2024)Yang, Yang, Hui, Zheng, Yu, Zhou, Li, Li, Liu, Huang,
  et~al.]{yang2024qwen2}
An Yang, Baosong Yang, Binyuan Hui, Bo Zheng, Bowen Yu, Chang Zhou, Chengpeng
  Li, Chengyuan Li, Dayiheng Liu, Fei Huang, et~al.
\newblock Qwen2 technical report.
\newblock \emph{arXiv preprint arXiv:2407.10671}, 2024.

\bibitem[Yu et~al.(2021{\natexlab{a}})Yu, Li, Koh, Zhang, Pang, Qin, Ku, Xu,
  Baldridge, and Wu]{VITVQGAN}
Jiahui Yu, Xin Li, Jing~Yu Koh, Han Zhang, Ruoming Pang, James Qin, Alexander
  Ku, Yuanzhong Xu, Jason Baldridge, and Yonghui Wu.
\newblock Vector-quantized image modeling with improved vqgan.
\newblock \emph{arXiv preprint arXiv:2110.04627}, 2021{\natexlab{a}}.

\bibitem[Yu et~al.(2021{\natexlab{b}})Yu, Li, Koh, Zhang, Pang, Qin, Ku, Xu,
  Baldridge, and Wu]{yu2021vector}
Jiahui Yu, Xin Li, Jing~Yu Koh, Han Zhang, Ruoming Pang, James Qin, Alexander
  Ku, Yuanzhong Xu, Jason Baldridge, and Yonghui Wu.
\newblock Vector-quantized image modeling with improved vqgan.
\newblock \emph{arXiv preprint arXiv:2110.04627}, 2021{\natexlab{b}}.

\bibitem[Yu et~al.(2022)Yu, Xu, Koh, Luong, Baid, Wang, Vasudevan, Ku, Yang,
  Ayan, et~al.]{yu2022scaling}
Jiahui Yu, Yuanzhong Xu, Jing~Yu Koh, Thang Luong, Gunjan Baid, Zirui Wang,
  Vijay Vasudevan, Alexander Ku, Yinfei Yang, Burcu~Karagol Ayan, et~al.
\newblock Scaling autoregressive models for content-rich text-to-image
  generation.
\newblock \emph{arXiv preprint arXiv:2206.10789}, 2\penalty0 (3):\penalty0 5,
  2022.

\bibitem[Yu et~al.(2023)Yu, Lezama, Gundavarapu, Versari, Sohn, Minnen, Cheng,
  Birodkar, Gupta, Gu, et~al.]{yu2023language}
Lijun Yu, Jos{\'e} Lezama, Nitesh~B Gundavarapu, Luca Versari, Kihyuk Sohn,
  David Minnen, Yong Cheng, Vighnesh Birodkar, Agrim Gupta, Xiuye Gu, et~al.
\newblock Language model beats diffusion--tokenizer is key to visual
  generation.
\newblock \emph{arXiv preprint arXiv:2310.05737}, 2023.

\bibitem[Zhao et~al.(2023)Zhao, Bai, Rao, Zhou, and Lu]{zhao2023unipc}
Wenliang Zhao, Lujia Bai, Yongming Rao, Jie Zhou, and Jiwen Lu.
\newblock Unipc: A unified predictor-corrector framework for fast sampling of
  diffusion models.
\newblock \emph{arXiv preprint arXiv:2302.04867}, 2023.

\bibitem[Zheng et~al.(2023)Zheng, Nie, Vahdat, Azizzadenesheli, and
  Anandkumar]{zheng2022fast}
Hongkai Zheng, Weili Nie, Arash Vahdat, Kamyar Azizzadenesheli, and Anima
  Anandkumar.
\newblock Fast sampling of diffusion models via operator learning.
\newblock In \emph{ICML}, 2023.

\bibitem[Zhu et~al.(2024)Zhu, Wei, Lu, and Chen]{zhu2024scaling}
Lei Zhu, Fangyun Wei, Yanye Lu, and Dong Chen.
\newblock Scaling the codebook size of vqgan to 100,000 with a utilization rate
  of 99\%.
\newblock \emph{arXiv preprint arXiv:2406.11837}, 2024.

\end{thebibliography}


\begin{thebibliography}{5}
\providecommand{\natexlab}[1]{#1}
\providecommand{\url}[1]{\texttt{#1}}
\expandafter\ifx\csname urlstyle\endcsname\relax
  \providecommand{\doi}[1]{doi: #1}\else
  \providecommand{\doi}{doi: \begingroup \urlstyle{rm}\Url}\fi

\bibitem[Frans et~al.(2024)Frans, Hafner, Levine, and Abbeel]{frans2024one}
Kevin Frans, Danijar Hafner, Sergey Levine, and Pieter Abbeel.
\newblock One step diffusion via shortcut models.
\newblock \emph{arXiv preprint arXiv:2410.12557}, 2024.

\bibitem[Kingma and Welling(2014)]{kingma2013auto}
Diederik~P Kingma and Max Welling.
\newblock Auto-encoding variational bayes.
\newblock In \emph{ICLR}, 2014.

\bibitem[Li et~al.(2025)Li, Tian, Li, Deng, and He]{li2025mar}
Tianhong Li, Yonglong Tian, He Li, Mingyang Deng, and Kaiming He.
\newblock Autoregressive image generation without vector quantization.
\newblock \emph{Advances in Neural Information Processing Systems},
  37:\penalty0 56424--56445, 2025.

\bibitem[Lipman et~al.(2022)Lipman, Chen, Ben-Hamu, Nickel, and
  Le]{lipman2022flow}
Yaron Lipman, Ricky~TQ Chen, Heli Ben-Hamu, Maximilian Nickel, and Matt Le.
\newblock Flow matching for generative modeling.
\newblock \emph{arXiv preprint arXiv:2210.02747}, 2022.

\bibitem[Sohn et~al.(2015)Sohn, Lee, and Yan]{sohn2015learning}
Kihyuk Sohn, Honglak Lee, and Xinchen Yan.
\newblock Learning structured output representation using deep conditional
  generative models.
\newblock \emph{Advances in neural information processing systems}, 28, 2015.

\end{thebibliography}
}

\clearpage
\setcounter{page}{1}
\maketitlesupplementary
\appendix

\section{Environment}
The software environment is based on PyTorch 2.4.1 with CUDA 11.8. All models are trained using $32\times$ NVIDIA V100 GPUs. For NVIDIA V100 GPUs, we utilize xFormers to accelerate attention computations. When measuring inference time on NVIDIA A100 GPUs, we use Flash Attention 2.5.9.

\section{More Implementation Details}

\noindent\textbf{Model Details.} Following MAR~\cite{li2025mar}, we set a minimum masking ratio of 0.7. As in MAR, we use \texttt{[CLS]} tokens to generate conditions. For FAR, we repeat the \texttt{[CLS]} token 64 times, while for FAR-Causal, we use a single \texttt{[CLS]} token. The network architecture details of FAR-B and FAR-L are available in Table~\ref{tab:mar_configs}.

\noindent\textbf{Shortcut Head Sampling Details.}
As shown in the main paper, our overall loss function combines the flow matching loss and the consistency loss. Our velocity field prediction network, $f_{\theta}$, can take $d=0$ as input, allowing it to function as a standard flow matching model. Empirically, we set $d = {1}/{\text{(number of sampling steps)}}$ when the number of sampling steps is 16 or fewer, and $d=0$ when it exceeds 16. The sampling process is formulated as:
\begin{align}
    \hat{\boldsymbol{h}}_0 &\sim \mathcal{N}(\boldsymbol{0}, \boldsymbol{I}), \\
    \hat{\boldsymbol{h}}_{t+\frac{1}{N}} &= \hat{\boldsymbol{h}}_{t} + \frac{1}{N} f_{\theta}(\hat{\boldsymbol{h}}_{t}, \boldsymbol{c}, d),
\end{align}
where $N$ represents the number of sampling steps, with $d = 1/N$ when $N \leq 16$, and $d=0$ when $N > 16$. The variable $t$ denotes the current sampling step, with $t \in \{0, {1}/{N}, {2}/{N}, \cdots, {(N-1)}/{N}\}$.

\noindent\textbf{Discussion.} Frans \textit{et al.}~\cite{frans2024one} discretely sample $t$ to ensure alignment between training and inference, selecting $d$ from a predefined list. They identify weight decay as a crucial factor in the training process. Our approach, which integrates a generative shortcut head, enables more stable training and yields greater performance improvements over flow matching~\cite{lipman2022flow}, even with fewer head sampling steps. Further exploration of the optimal distribution of step size $d$ remains a promising direction for future research.

\section{More Experiments}

\noindent\textbf{Ablation Study on FAR-B-Causal.} 
FAR-B-Causal consists of a causal Transformer and a FAR head with a shortcut implementation. We compare FAR-B-Causal with a variant in which the head network is replaced by the diffusion head used in MAR~\cite{li2025mar}, referred to as MAR-B-Causal.

Table~\ref{tab:supp-ar-diffusion-vs-shortcut} presents the performance comparison between FAR-B-Causal and MAR-B-Causal. Both models are trained for 400 epochs, and share the same architecture—a 24-layer causal Transformer and a 6-layer MLP head—differing only in their approach to denoising noisy tokens. FAR-B-Causal achieves the best overall performance, attaining an FID of 5.67 with 50 sampling steps, outperforming MAR-B-Causal, which reaches an FID of 6.80 even with 100 steps. Notably, with just 2 sampling steps, FAR-B-Causal achieves an FID of 7.86 while obtaining the highest Inception Score (260.24) among all configurations. These results demonstrate the efficiency of our approach, delivering strong performance with significantly reduced computational overhead during inference.

\begin{table}[!t]
    \centering
    \begin{tabular}{cccc}
    \toprule
      Model   & Denoising Steps & FID$\downarrow$ & IS$\uparrow$ \\
    \midrule
      MAR-B-Causal  & 100 & 6.80 & 224.31 \\
      FAR-B-Causal  & 2 & 7.86 & 260.24 \\
      FAR-B-Causal  & 50 & 5.67 & 226.18 \\
    \bottomrule 
    \end{tabular}
    \caption{Performance comparison between FAR-B-Causal and MAR-B-Causal. Both models are trained for 400 epochs and share the same architecture, differing only in their denoising approach for noisy tokens.}
    \label{tab:supp-ar-diffusion-vs-shortcut}
\end{table}

\begin{table}[!t]
\centering
\begin{tabular}{ccc}
\toprule
KL Weight & FID$\downarrow$ & IS$\uparrow$ \\
\midrule
0.1 & 4.25 & 262.26 \\
0.01 & 3.33 & 281.03 \\
0.001 & 3.73 & 277.92 \\
0.0005 & 3.94 & 283.46 \\
0.0002 & 4.20 & 291.84 \\
0.0001 & 4.48 & 302.68 \\
\bottomrule
\end{tabular}
\caption{Study on FAR-B variants with a C-VAE head, each using a different KL weight.}
\label{tab:kl_weight_comparison}
\end{table}

\noindent\textbf{C-VAE as One-step Generative Head.} In the main paper, we demonstrate the superiority of our shortcut head over the diffusion head and flow matching head. Here, we explore an alternative head network based on a conditional VAE (C-VAE)~\cite{kingma2013auto,sohn2015learning} implementation. 

\begin{table*}[h]
    \centering
    \begin{tabular}{lcccccc}
    \toprule
    \multirow{2}{*}{Model} & Encoder & Encoder & Encoder & Decoder & Decoder & Decoder \\
     & Embed. Dim. & Depth & Heads & Embed. Dim. & Depth & Heads \\
    \midrule
    FAR-B & 768 & 12 & 12 & 768 & 12 & 12 \\
    FAR-L & 1024 & 16 & 16 & 1024 & 16 & 16 \\
    \bottomrule
    \end{tabular}
    \caption{FAR model architecture configurations. ``Embed. Dim.'': embedding dimension.}
    \label{tab:mar_configs}
\end{table*}

Specifically, the C-VAE head consists of a 3-layer MLP encoder and a 3-layer MLP decoder. The encoder takes a ground-truth token $\boldsymbol{z}$ as input and integrates the condition $\boldsymbol{c}$ using the Adaptive Layer Normalization operation, similar to the condition injection method in our shortcut head. The latent representations generated by the encoder are then passed to the decoder to reconstruct $\boldsymbol{z}$, conditioned on $\boldsymbol{c}$. The condition injection in the decoder follows the same approach as in the encoder. To enable sampling during inference, KL divergence regularization is applied to enforce the latent space's proximity to a standard Gaussian distribution. During inference, the encoder is discarded. A Gaussian noise sample is drawn randomly and combined with the given condition $\boldsymbol{c}$ to generate image tokens via the decoder.

In Table~\ref{tab:kl_weight_comparison}, we examine variants of FAR-B with a C-VAE head, each using a different KL weight. With a KL weight of 0.01, FAR-B with a C-VAE head achieves the best FID (3.33) among all variants. Although its performance is slightly lower than FAR-B with a shortcut head, its computational cost is significantly lower, as the C-VAE head only needs to be called once per image token generation.

\end{document}